\documentclass[conference]{IEEEtran}
\IEEEoverridecommandlockouts
\usepackage{cite}
\usepackage{amsmath,amssymb,amsfonts}

\usepackage{graphicx}
 
\usepackage{float} 
\usepackage{textcomp}

\usepackage{array}
\usepackage{multirow}
\usepackage{arydshln}

\usepackage{graphicx,subcaption}
\usepackage{wrapfig}
\usepackage{xcolor}
\usepackage{tikz,pgfplots}
\usetikzlibrary{arrows}

\usepackage[acronym]{glossaries}

\usepackage{algorithm}
\usepackage{algpseudocode}
\usepackage{algorithmicx}

\usepackage{caption}
\usepackage[numbers]{natbib}

\usepackage{hyperref} 
\hypersetup{
    linkbordercolor={0 0 1}
}

\usepackage{xcolor}
\def\BibTeX{{\rm B\kern-.05em{\sc i\kern-.025em b}\kern-.08em
    T\kern-.1667em\lower.7ex\hbox{E}\kern-.125emX}}
\begin{document}

\title{Design and Development of a Research Oriented Low Cost Robotics Platform with a Novel Dynamic Global Path Planning Approach\\}

\author{Shalutha Rajapakshe\IEEEauthorrefmark{2}\IEEEauthorrefmark{1}  
        and Ramith Hettiarachchi\IEEEauthorrefmark{2}\IEEEauthorrefmark{1}\\
        \IEEEauthorrefmark{2}Robotics and Autonomous Systems Group, CSIRO, Pullenvale, QLD 4069, Australia\\
        \IEEEauthorrefmark{1}Department of Electronic and Telecommunication Engineering, University of Moratuwa, Sri Lanka \\
        e-mail : shalutha321@gmail.com, ramithuh@live.com\vspace{-0.35cm}
        }

\maketitle

\begin{abstract}

Autonomous navigation systems based on computer vision sensors often require sophisticated robotics platforms which are very expensive. This poses a barrier for the implementation and testing of complex localization, mapping, and navigation algorithms that are vital in robotics applications. 
Addressing this issue, in this work, Robot Operating System (ROS) supported mobile robotics platforms are compared and an end-to-end implementation of an autonomous navigation system based on a low-cost educational robotics platform, AlphaBot2 is presented, while integrating the Intel RealSense D435 camera. Furthermore, a novel approach to implement dynamic path planners as global path planners in the ROS framework is presented. We evaluate the performance of this approach and highlight the improvements that could be achieved through a dynamic global path planner. 
This low-cost modified AlphaBot2 robotics platform along with the proposed dynamic global path planning approach will be useful for researchers and students for getting hands-on experience with computer vision-based navigation systems.

\end{abstract}

\section{Introduction}
Autonomous mobile robots have the ability to understand their surrounding environments and perform path-planning to complete navigation tasks.
Usually, for these autonomous mobile robots, robotics platforms with high-performance compute devices and sophisticated sensors (LIDAR, RADAR, and vision systems) are required for algorithms such as simultaneous localization and mapping (SLAM), which enable robots to build maps of their surroundings \cite{Ismail2019SurveyHardware}. Therefore, high costs associated with such robotics platforms hinder learning and implementation of autonomous navigation systems, which impedes research opportunities especially for researchers getting started in the robotics field.

Path planning is a vital component of an autonomous robot. The navigation task is achieved by the use of global and local path planners which focuses on finding the optimal or sub-optimal path, and obstacle avoidance respectively. However, the performance evaluations based on the trade-off between these two types of planners have not been explored much in the community. The use of a local planner to avoid dynamic obstacles does not necessarily generate optimized paths considering the global context. An example for this would be the Time Elastic Bands method used as the local planner, \cite{Marin-Plaza2018GlobalVehicles} where the dynamic obstacles' directions influence the path planning, leading to inefficient plans. Furthermore, relying solely on local planners to avoid obstacles may lead to getting trapped in local minima \cite{Cai2020MobileSurvey}. Recent work by Niu \textit{et al}. has presented 
a method for global dynamic path planning by the use of the dynamic window method along with an improved A* algorithm \cite{Niu2021ResearchAlgorithm}. It enables avoiding dynamic obstacles while moving towards the end goal using the shortest length. However, this method is limited to the use of the static planner A* and considers the global context only based on the initially planned global path.

\begin{figure}[H]
    \centering
      \includegraphics[width=0.15\textwidth]{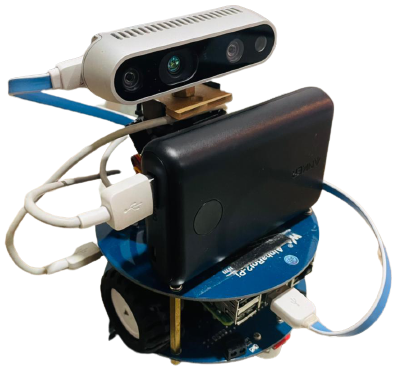}
    \caption{Modified AlphaBot2 robotics platform}
    \label{fig:alphabot}
\end{figure}

AlphaBot2 \cite{waveshare_alphabot2} by Waveshare Electronics is a low-cost educational robotics platform. Due to performance limitations of its compute device, in our work, we utilize this robotics platform along with a Raspberry Pi 4 device having 8GB of random access memory (RAM). The sensors and actuators contained in the modified AlphaBot2 platform are summarized in Table \ref{tab:comparison}. Even though the AlphaBot2 robotics platform uses a Raspberry Pi compute device, it does not have ROS pre-installed. Furthermore, as of now, there are no official ROS packages developed for AlphaBot2.

\begin{table*}[ht]
\caption{ROS supported mobile robot platforms}
\label{tab:comparison}
\resizebox{\textwidth}{!}{
\begin{tabular}{|l|c|c|c|c|c|c|}
\hline
                                                                              & \textbf{\begin{tabular}[c]{@{}c@{}}TurtleBot3\\ (Burger)\end{tabular}}    & 
                                                                              
                                                                        \textbf{ROSbot 2.0}                                                                                                       & \textbf{\begin{tabular}[c]{@{}c@{}}Jet Racer\end{tabular}}                                                                                                            & \textbf{JetBot AI Kit}                                                                           & \textbf{\begin{tabular}[c]{@{}c@{}}DuckieBot\\  (DB19)     \end{tabular}}                                          & \textbf{\begin{tabular}[c]{@{}c@{}}AlphaBot2 \\ +\\ Intel RealSense \\  + \\ AnkerPowerCore II\end{tabular}}                                                         \\ \hline
\textbf{\begin{tabular}[c]{@{}l@{}}Single-Board \\ Computer/CPU\end{tabular}} & Raspberry Pi 3                                                            & {\color[HTML]{000000} \begin{tabular}[c]{@{}c@{}}Asus Tinker Board \\ with  Rockchip \\ RK3288\end{tabular}}              & \begin{tabular}[c]{@{}c@{}}Nvidia\\ Jetson Nano\end{tabular}                                                                        & \begin{tabular}[c]{@{}c@{}}Nvidia\\ Jetson Nano\end{tabular}                               & \begin{tabular}[c]{@{}c@{}}Raspberry Pi 3\\ Model 3\end{tabular}                                              & \begin{tabular}[c]{@{}c@{}}Raspberry Pi 4\\ Model B\end{tabular}                                                                                                   \\ \hline
\textbf{GPU}                                                                  & Broadcom VideoCore IV                                                                       & \begin{tabular}[c]{@{}c@{}}ARM Mali-T764 \\ MP2\end{tabular}                                                              & \begin{tabular}[c]{@{}c@{}}128-core Maxwell \\ GPU\end{tabular}                                                                     & \begin{tabular}[c]{@{}c@{}}128-core Maxwell \\ GPU\end{tabular}                            & Broadcom VideoCore IV                                                                                                           & Broadcom VideoCore VI                                                                                                                                                                \\ \hline
\textbf{Actuators}                                                            & \begin{tabular}[c]{@{}c@{}}DYNAMIXEL\\ XL430\end{tabular}                 & \begin{tabular}[c]{@{}c@{}}Xinhe Motor \\ XH-25D\end{tabular}                                                             & \begin{tabular}[c]{@{}c@{}}37-520 DC \\ gearmotor\\ Reduction rate 1:10\\ Idlespeed 740RPM.\\ MG996R, \\ 9kg/cm torque\end{tabular} & \begin{tabular}[c]{@{}c@{}}TT motor\\ Reduction rate 1:48\\ Idle speed 240RPM\end{tabular} & \begin{tabular}[c]{@{}c@{}}Two DC motors \\ with (Hall effect \\ sensor based) \\ wheel encoders\end{tabular} & \begin{tabular}[c]{@{}c@{}}N20 micro \\ gear motor \\ reduction rate 1:30, \\ 6V/600RPM\end{tabular}                                                               \\ \hline
\textbf{Camera}                                                               & N/A                                                                       & \begin{tabular}[c]{@{}c@{}}Orbbec Astra \\ RGBD \\ camera\end{tabular}                                                    & \begin{tabular}[c]{@{}c@{}}Sony IMX219, \\ 8MP, 160 FOV \\ wide angle camera\end{tabular}                                           & \begin{tabular}[c]{@{}c@{}}IMX219 8MP, \\ 160 FOV \\ wide angle camera\end{tabular}        & FBA\_RPi Camera (F)                                                                                           & \begin{tabular}[c]{@{}c@{}}Intel RealSense \\ D435\end{tabular}                                                                                                    \\ \hline
\textbf{\begin{tabular}[c]{@{}l@{}}Additional\\ Sensors\end{tabular}}         & \begin{tabular}[c]{@{}c@{}}360 LIDAR\\ (HLS-LFCD2)\end{tabular}           & \begin{tabular}[c]{@{}c@{}}RPLIDAR A2 \\ laser scanner,\\ MPU 9250, \\ VL53L0X,\\ 4 x quadrature \\ encoders\end{tabular} & N/A                                                                                                                                 & N/A                                                                                        & N/A                                                                                                           & \begin{tabular}[c]{@{}c@{}}Reflective infrared \\ photoelectric \\ sensor (ST188)\\ \\ Reflective infrared \\ photoelectric \\ sensor \\ (ITR20001/T)\end{tabular} \\ \hline
\textbf{OS}                                                                   & \begin{tabular}[c]{@{}c@{}}TurtleBot3 \\ Raspbian\end{tabular}            & \begin{tabular}[c]{@{}c@{}}Ubuntu-based OS, \\ customized to use \\ ROS.\end{tabular}                                     & Ubuntu 18.04 LTS                                                                                                                    & Ubuntu 18.04 LTS                                                                           & Ubuntu-based OS                                                                                               & Ubuntu 20.04 LTS                                                                                                                                                       \\ \hline
\textbf{Wireless}                                                             & \begin{tabular}[c]{@{}c@{}}2.4GHz \\ 802.11n wireless\end{tabular}        & \begin{tabular}[c]{@{}c@{}}802.11 b/g/n with \\ upgradable \\ IPEX antenna\end{tabular}                                   & \begin{tabular}[c]{@{}c@{}}2.4GHz / 5GHz\\  dual-band WiFi, \\ Bluetooth 4.2\end{tabular}                                           & \begin{tabular}[c]{@{}c@{}}2.4GHz / 5GHz\\  dual-band WiFi, \\ Bluetooth 4.2\end{tabular}  & \begin{tabular}[c]{@{}c@{}}Edimax AC1200 \\ EW-7822ULC \\ 5 GHz wireless \\ adapter\end{tabular}              & \begin{tabular}[c]{@{}c@{}}2.4 GHz / 5.0 GHz \\ IEEE 802.11ac \\ wireless, \\ Bluetooth 5.0, \\ BLE\end{tabular}                                                   \\ \hline
\textbf{Power Supply}                                                         & \begin{tabular}[c]{@{}c@{}}Li-Po Battery \\ 11.1V \\ 1800mAh\end{tabular} & \begin{tabular}[c]{@{}c@{}}Li-on batteries: \\ 3 x 3500 mAh\end{tabular}                                                  & \begin{tabular}[c]{@{}c@{}}12.6V, \\ 18650 battery × 3\end{tabular}                                                                 & \begin{tabular}[c]{@{}c@{}}12.6V, \\ 18650 battery × 3\end{tabular}                        & \begin{tabular}[c]{@{}c@{}}RAVPower Portable \\ Charger 10400mAh\end{tabular}                                 & \begin{tabular}[c]{@{}c@{}}Anker PowerCore \\ II \\ 10000 mAh\end{tabular}                                                                                         \\ \hline
\textbf{Total Cost}                                                                & \$612                                                            & \$1916                                                                                                        & \$396                                                                                                                           & \$302                                                                                 & \$302                                                                                                       & \$371                                                                                                                                                         \\ \hline
\end{tabular}}
\end{table*}

Robotics integrated STEM education can boost the students’ interests and career orientation to a higher level while enhancing their problem solving and mathematical skills. This has been proven by chen \textit{et al.} through a statistical analysis considering a student population  \cite{Chen2018TheOrientation}. This marks the importance of having access to robotics platforms which can teach wide range of robotics applications. A recent publication has reviewed robotics platforms for educational purposes and compares AlphaBot2 with other robotics platforms \cite{Sataloff}. They state the limitations in terms of vision and mapping aspects of AlphaBot2 and argues that improving those aspects could lead to a very beneficial robotics platform. There are several applications of AlphaBot2 in the literature \cite{Ferreira2020,Rafael2020a} which include educational robotics projects as well. A Gazebo-based simulation framework has been introduced by Rafael \textit{et al.} which is helpful for students to control AlphaBot2 in both real and simulated environments \cite{Rafael2020a}. A feature-rich, inexpensive robot, based on AlphaBot2 is introduced by Ferreira \textit{et al.} where they produce an omnidirectional vision system based on a Raspberry Pi Camera Module \cite{Ferreira2020}. Even though there are some applications of AlphaBot2 in the areas of internet of things (IoT) \cite{Kuo2018} and computer vision \cite{Ferreira2020}, there were no attempts in applying complex SLAM algorithms in such a compact and cost-effective mobile robot. Therefore, we explore the possibility of applying such complex localization, mapping, and navigation algorithms on AlphaBot2 by leveraging the robust RGB-D data from the Intel RealSense D435 camera. In our work, the main contributions are as follows:

\begin{itemize}
    \item Present a comprehensive comparison of ROS-supported off-the-shelf mobile robotics platforms.
    
    \item Enable autonomous navigation on AlphaBot2 by performing hardware and software modifications.

    \item Present a novel dynamic global path planning approach to minimize computational costs.   
 
     \item Develop and evaluate an end-to-end setup on the modified AlphaBot2 platform to perform localization, mapping, and navigation, while using D* lite algorithm as the dynamic global path planner through the proposed method.
    
\end{itemize}

\section{Methodology}

\vspace{-0.12cm}

This section is divided into six subsections. Section \ref{sec:setting_up_alphabot2} describes the software and hardware setup of AlphaBot2 and the used computer vision algorithms are discussed in section \ref{sec:computer_vision}. Section \ref{sec:dynamic_path_planning} and \ref{sec:enabling_dynamic} discuss the current limitations in implementing dynamic path planners as global planners in ROS and our novel approach respectively.
Our system was tested using simulations as described in section \ref{sec:simulation}, and the actual implementation details are included in section \ref{sec:implementation_on_modified_alphabot2}.

We perform a comprehensive comparison of various ROS supported mobile robot platforms which are suitable for localization, mapping and navigation tasks (Table \ref{tab:comparison}). While Turtlebot3 and ROSbot 2.0 are highly suitable for research tasks considering their highly capable sensors and actuators, their high cost factor limits researchers from accessing them. Meanwhile the most cost-effective robotic platforms which supports ROS can be identified as Jet Racer, JetBot AI Kit, DuckieBot, and AlphaBot2. Yet, these platforms do not have sophisticated sensors to perform complex visual-SLAM and navigation tasks. Therefore, considering the trade-off between the cost factor and capabilities, we identified that AlphaBot2 and the Intel RealSense D435\cite{web_realsense} camera powered by Anker PowerCore II\cite{web_anker} would be an ideal combination.

\subsection{Setting up AlphaBot2}
\label{sec:setting_up_alphabot2}

\begin{figure}[H]
    \centering
      \includegraphics[width=0.36\textwidth]{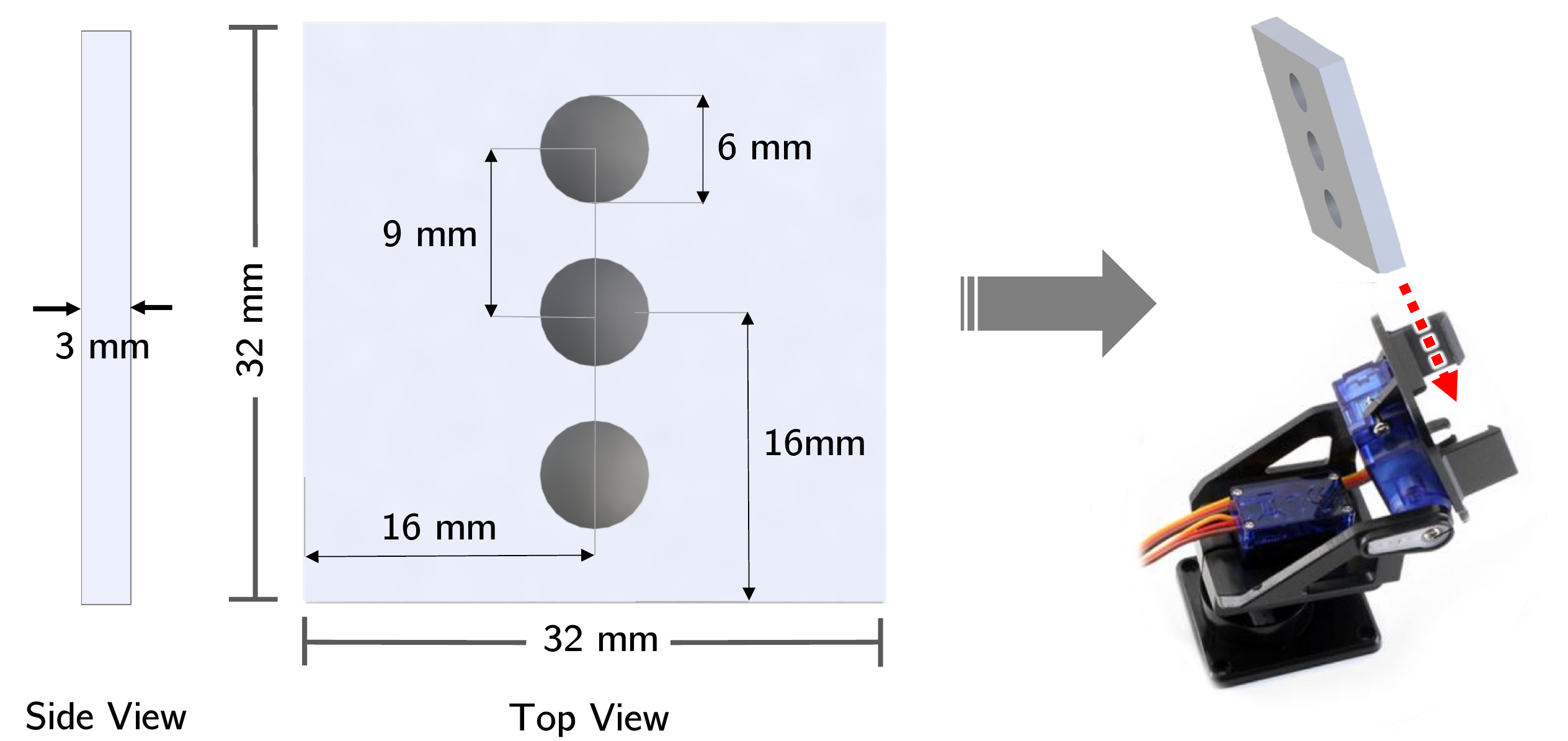}
    \caption{Realsense camera adapter and installation}
    \label{fig:adapter}
\end{figure}

Initially, Ubuntu Server 20.04 was installed on the Raspberry Pi 4 device and then ROS Noetic version was installed. To perform SLAM with a sophisticated camera which gives RGB-D data, we used an Intel RealSense D435 camera instead of the default RaspberryPi Camera. This camera is used as an add-on to the robot base through a hardware adapter we designed shown in Figure~\ref{fig:adapter}. This makes it possible to simply attach the camera without major hardware modifications as shown in Figure~\ref{fig:adapter}, with minimum effect on the weight factor. Due to the power consideration of the external camera and the capacity limitation of pre-installed batteries, we installed an Anker PowerCore II powerbank (capacity: 10000mAh, output: 5-9V/2A). This enabled an extended usage time of the robot ranging between 1.5 - 2 hours while working under full capacity. The modified AlphaBot2 robot is shown in Figure \ref{fig:alphabot}.

\subsection{Computer Vision Algorithms}
\label{sec:computer_vision}

Real-Time Appearance-Based Mapping (RTAB-Map) is a ROS compatible, graph-based visual-SLAM approach consisting of a feature based visual odometry front-end and a pose graph optimization back-end. It supports RGB-D and Stereo modes for visual SLAM and can fuse input streams from IMU, LIDAR, and robot odometry. Visual odometry is a method that determines the odometry information of a robot by analyzing a sequence of camera images. This can provide accurate and reliable localization estimates compared to wheel odometry \cite{Yousif2015} which uses encoder information. In our work, RTAB-Map is used on the incoming RGB-D data from the Intel RealSense D435 camera for localization and mapping purposes.

\subsection{Dynamic Path Planning}
\label{sec:dynamic_path_planning}

Incremental heuristic search methods reuse information from previous searches and uses heuristic knowledge as approximations for faster re-planning. D* lite is an example for such type of algorithm. Even though it has the same navigation strategy as Dynamic A* (D*) algorithm \cite{Stentz1994}, it is substantially shorter and algorithmically simpler than D* and simplifies the analysis of the program flow. Therefore, we have selected D* lite as a potential path planning algorithm to be used in our application.

In the ROS navigation stack, \textit{move\_base} node is responsible for moving a robot to a given target location. 
To achieve this task, \textit{move\_base} enables customization of global and local path planners as plugins to ROS. These planners make use of local and global costmaps. Global planner considers the map of the whole environment and calculates a high level plan to a given destination, while the local planner takes dynamic obstacles into account and recalculates paths at a user-defined rate. This recalculating process only considers the surrounding information of the robot which is captured by its sensors.

Dynamic path planning is commonly performed in ROS by initially calculating a path using a static planner such as A*, Dijkstra's algorithms, which acts as the global planner followed by a local planner which will handle the trajectory planning and obstacle avoidance. Example local planner approaches include Time Elastic Band (TEB), and Dynamic Window Approach (DWA).
Since the local planner only contains information about the robot's immediate surrounding, the new trajectories it generates to avoid obstacles might not be the best path considering the global context. One option to overcome this issue would be to call the global planner at a defined frequency to recalculate the best path considering the new global map. For this purpose, the ROS ecosystem provides a parameter named $planner\_frequency$ in its navigation stack. However, this method will recalculate the global path from its current position to goal position even if there are no changes in the surrounding environment leading to inefficient computations which consume time and energy. 

To solve these issues, we provide a novel algorithm which uses the global planner to recalculate the path only if there are changes detected in the surrounding environment, when a new plan is requested at the planner frequency rate. Furthermore, this algorithm enables the usage of the re-planning aspect of dynamic path planning algorithms in the ROS environment. Currently, this has not been possible in the default ROS global path planner setup. The functional aspect of our novel algorithm is described in detail in section \ref{sec:enabling_dynamic}.

\subsection{Enabling Dynamic Global Path Planning in ROS}
\label{sec:enabling_dynamic}

Initially, a path planner service server is created with D* lite as the global planner. Then the planner is registered as a plugin and a client is created to request the path at the defined planner frequency. Our novel algorithm which is part of the path planner service server is listed in Algorithm \ref{alg1}.

In the path planner service server, global variables will be initialized to store the previous costmaps, plan, start, and goal positions. Once a request is made to the server, the function $make\_plan$ will be executed. This function will extract the updated cost map, current start, and goal positions. Then it checks whether the goal position has changed. If so, the global planner algorithm will be executed to generate a new path. If the goal position has not changed, the \textsc{calculatePlan} function will be called to check whether re-planning is needed. 

The \textsc{calculatePlan} function, will first calculate the costmap difference between the previous and the current costmaps. In this calculation, a threshold can be introduced as a noise suppression method since there could be minor changes in the costmap even without any obstacle changes. After computing the costmap difference, and if differences exist, it will determine the changed nodes and enter the re-planning state of the dynamic path planner. This process first updates changed nodes and then reuses information from the previous search which was used to generate the global plan. If no changes are detected in the costmap, it iterates through the previously planned global path's positions and identifies the closest position with respect to the robot's current location. After determining the closest position, it returns the rest of the positions to be followed towards the goal position from the previously calculated global path. This will reduce unnecessary path planning calculations during each time a plan is requested until the goal is reached and it leads to energy savings of the robot. The benefits of our approach can be summarized as follows,
\begin{itemize}
    \item Ability to provide an optimized path considering the global map's context.
    \item Re-planning will only be performed if the robot detects any change in the surrounding environment.
    
    
   %
    \item Ability to use algorithms with similar dynamic path planning capabilities.
\end{itemize}

\subsection{Simulation}
\label{sec:simulation}

We performed simulations in various map conditions on a custom built simulator as well as in the Gazebo simulator in ROS that consists of a physics engine. The custom-built simulator has the capability to identify cells that are not traversable due to a given robot size by marking virtual obstacles according to the given robot size, providing more similar real-world conditions. This will prevent the simulated robot from traversing through sharp corners of obstacles and traversing in between two obstacles.

\begin{algorithm}[t!]
\caption{Re-planning Algorithm}
\label{alg1}
\small
\begin{algorithmic}[1]

  \Statex
  \vspace{-0.1cm}
  \Procedure{calculatePlan}{$ $}
    \State $minDist = Infinity;$
    \State $costMapDiff$ $=$ $0;$
    \State $costMapDiff$ $ =$ $costDiff(costmap,$
    \State \hspace{4cm} $previousCostMap);$

    \If {$costMapDiff$ $>$ $0$}
        \State $changedNodes$ $=$ $findChangedNodes(costmap$
        \State \hspace{3.5cm} $ , previousCostMap);$
        
        \For{$node$ $in$ $changedNodes$}
        \State $updateNode(node);$
        \EndFor
        
        \State $plan$ $=$ $Replan();$
        \State $return$ $plan;$

    \Else
         \For{$pos$ $in$ $previousPlan$}
            
            \State $dist$ $=$ $euclidean\_dist(previousPlan[pos]$
            \State \hspace{4cm}$,currentStart);$
            
           
            \If {$dist$ $< $ $minDist;$}
            \State $minDist$ $=$ $dist;$
            \State $minDistIndex$ $=$ $pos;$
        
            \EndIf
        \EndFor   
         
        \State $return$ $previousPlan[minDistIndex::];$ 
    \EndIf
    
  \EndProcedure
  \Statex
  \vspace{-0.32cm}
  \Procedure{make\_plan}{$req$}
    \State $currentStart = req.start ;$
    \State $goal = req.goal;$
    \State $costmap = req.costmap;$
    \If {$previousGoal$ $!=$ $goal$}
        \State $plan$ $=$ $DStarLite(currentStart,goal,costmap);$
        \State $previousPlan$ $=$ $plan;$
    \Else
        \State $plan$ $=$ $CALCULATEPLAN(previousPlan$
        \State \hspace{2.5cm} $,currentStart,goal,costmap);$
    \EndIf
    
    \State $previousCostMap$ $=$ $costmap;$
    \State $response.plan$ $=$ $plan;$
    \State $return \ response; $

    
  \EndProcedure
\end{algorithmic}
\end{algorithm}

Using the novel dynamic global path planner method described in section \ref{sec:enabling_dynamic}, Turtlebot3-Burger robot model was used to benchmark path planning algorithms in Gazebo.
$turtlebot3\_navigation$ package was used and its $move\_base$ package's BaseGlobalPlanner parameter was modified by adding the newly created path planner plugin and $DWA\_local\_planner$ was added as the local planner. A launch file was then written with the required nodes to test the full setup on Turtlebot3 and the $planner\_frequency$ parameter was modified to request a new plan 2 times within one second from the D* lite algorithm. The evaluation procedure and the performance of the novel dynamic global path planner method is discussed in the results section.

\vspace{-0.09cm}
\subsection{Implementation on Modified AlphaBot2}
\label{sec:implementation_on_modified_alphabot2}
\begin{figure}
\begin{minipage}[b]{\linewidth}
\centering
\centerline{\includegraphics[width=0.95\columnwidth]{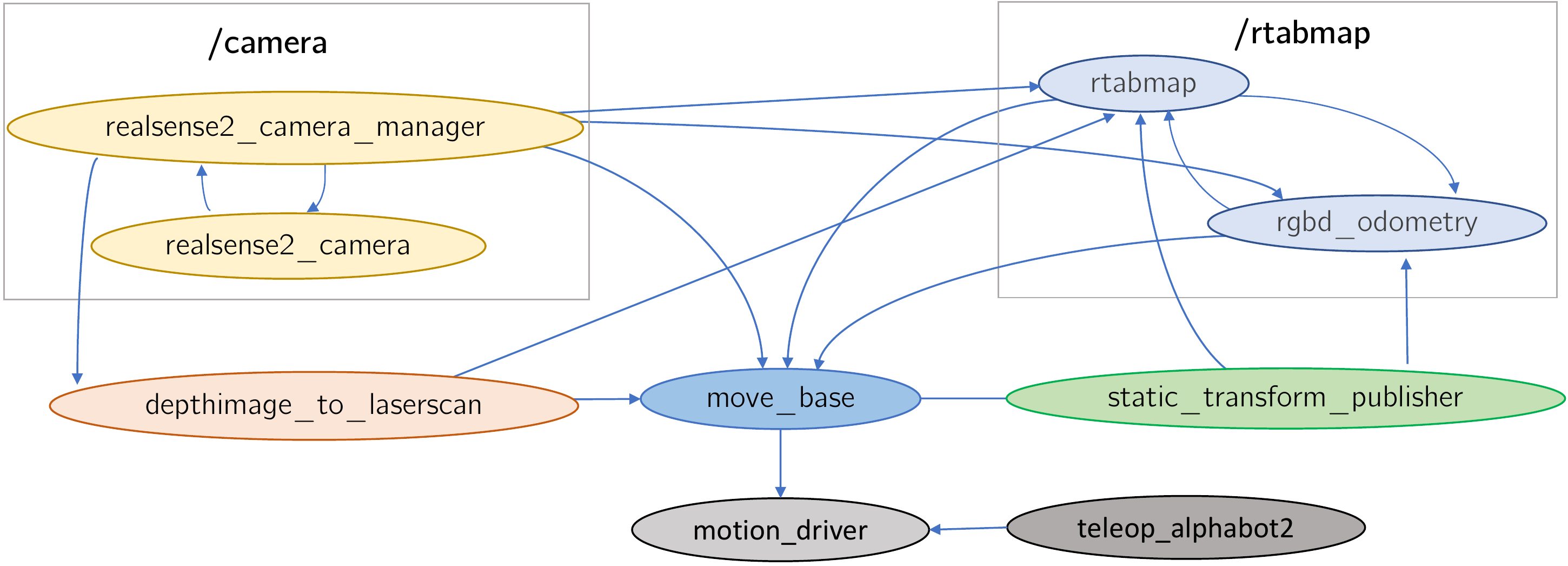}}
\caption{Interconnection of ROS nodes of the system}
\label{fig:ros_nodes}
\end{minipage}
\end{figure}

After performing simulations of the path planner, we implemented it on the actual AlphaBot2 robot. The interconnection of ROS nodes depicted in Figure \ref{fig:ros_nodes} summarizes all the necessary nodes that are required for the mapping and autonomous navigation tasks. Since AlphaBot2 does not have sensors such as motor encoders, inertial measurement units (IMU) to provide control feedback, we have implemented a motor controller node, $motion\_driver$ as a controller which provides optimized step movements for linear and angular velocity commands.

\vspace{-0.20cm}
\section{Results}

We evaluate our implementation based on two aspects: 1) The effectiveness of our modified AlphaBot2 platform for mapping and autonomous navigation tasks. 2) Performance of the novel dynamic replanning approach in simulation to solely study the proposed algorithm.

\vspace{-0.1cm}
\subsection{Effectiveness of the Modified AlphaBot2 Platform}
\begin{figure}[H]
    \centering

    \begin{subfigure}[b]{0.42\columnwidth}
        \centering
        \includegraphics[width=0.65\linewidth]{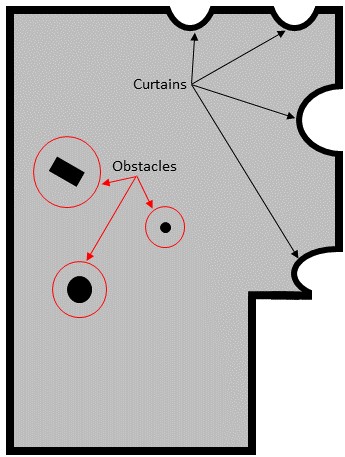}
        \caption{Groundtruth map}
        \label{fig:B}
    \end{subfigure}
    \begin{subfigure}[b]{0.48\columnwidth}
        \centering
        \includegraphics[width=0.65\linewidth]{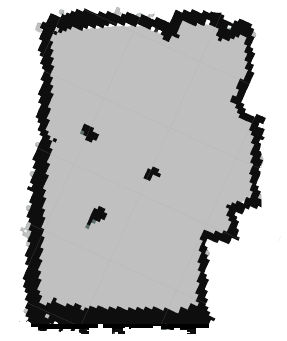}
        \caption{Generated map}
        \label{fig:B}
    \end{subfigure}
    \caption{Comparison of the generated and groundtruth maps}
    \label{fig:maps_}
\end{figure}

We performed several mapping sessions in an indoor environment to investigate the quality of the generated maps while navigating using our controller. A generated map along with its groundtruth map is shown in Figure \ref{fig:maps_} for comparison.

\begin{figure}[H]
    \centering

    \begin{subfigure}[b]{0.3\columnwidth}
        \centering
        \includegraphics[width=\linewidth]{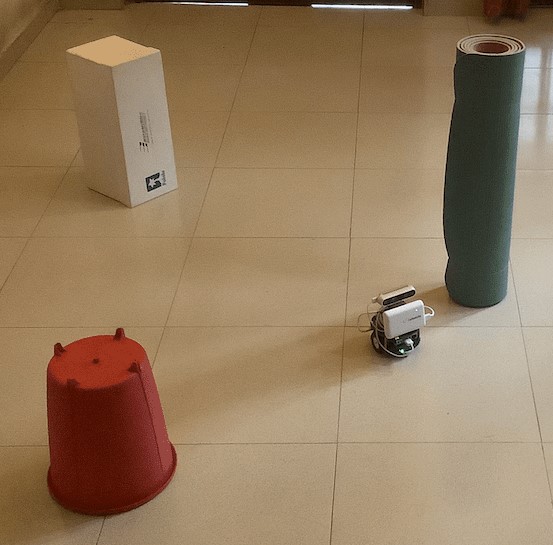}
        \label{fig:B}
    \end{subfigure}
    \begin{subfigure}[b]{0.235\columnwidth}
        \centering
        \includegraphics[width=\linewidth]{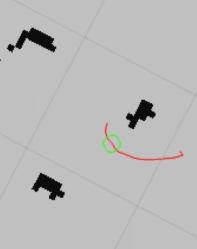}
        \label{fig:B}
    \end{subfigure}
    \vspace{-0.25cm}
    \caption{An instance of AlphaBot2 while it is following a generated path}
    \label{fig:navigation_results}
\end{figure}

After generating the map, we perform autonomous navigation based on our novel replanning algorithm, which makes use of the D* lite algorithm as the dynamic global path planner in the presence of various obstacles. The path calculated from the D* lite algorithm and an instance of AlphaBot2 following the generated path in a real scenario is shown in Figure \ref{fig:navigation_results}. 
We measure the CPU and RAM utilization on Raspberry Pi 4 during mapping and localization scenarios. It was observed that during the mapping mode, the average CPU utilization was approximately 84\%, while in the localization mode it varied between 88.5\% - 92.6\%. In both of the scenarios RAM utilization was approximately 21\%. To measure the average CPU and RAM utilization, the Linux tools ``s-tui" and ``htop" were used respectively for 10 minutes. This marks the feasibility of our proposed platform in complex robotics applications. Furthermore, due to the small form factor of the proposed robot platform, it could be ideal for applications which require tasks to be performed in a confined environment.

\vspace{-0.2cm}

\vspace{-0.4cm}

\vspace{-0.55cm}

\subsection{Evaluation of the Dynamic Global Path Planning Method }

\begin{figure}[H]
\centering
    \begin{subfigure}[b]{0.37\columnwidth}
        \centering
        \includegraphics[width=0.85\linewidth]{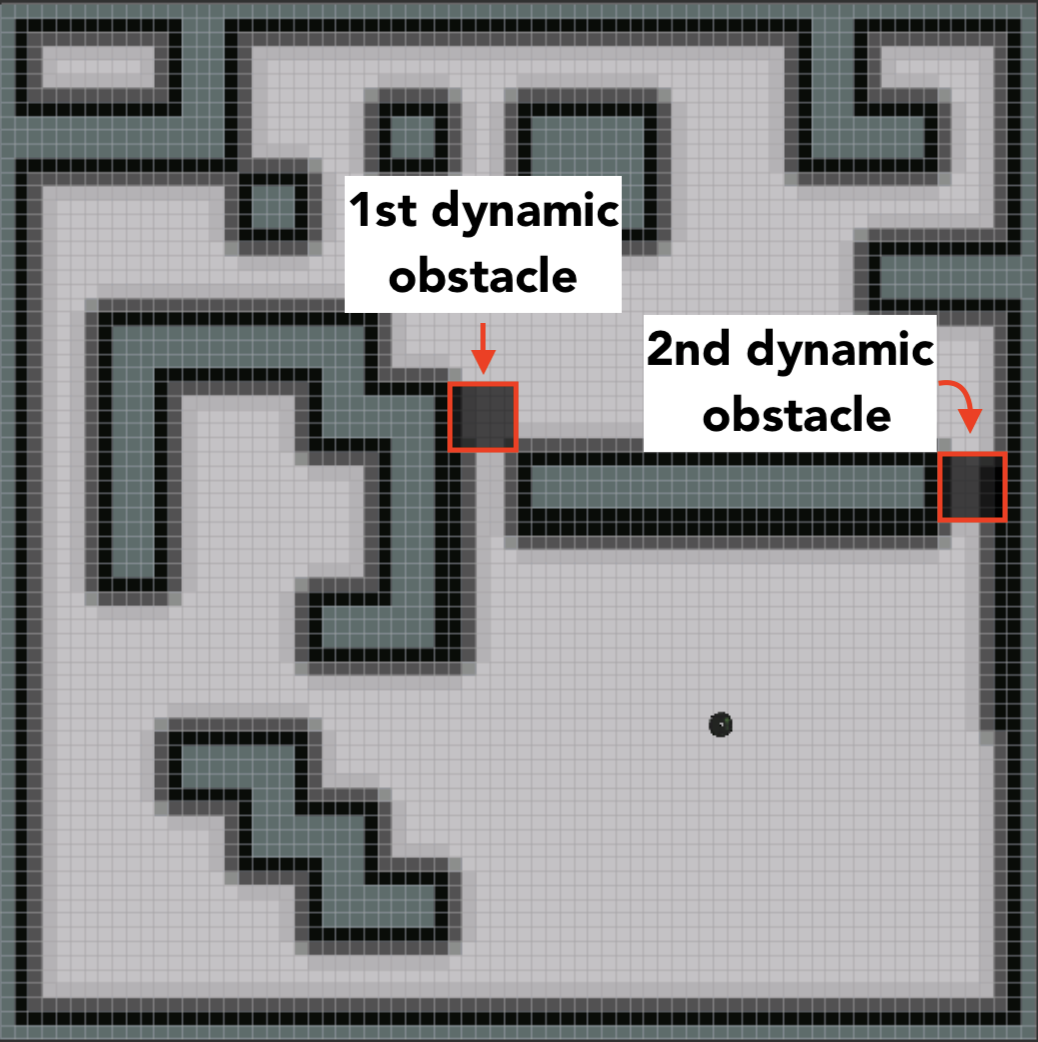}
        \caption{Obstacle locations}
        \label{fig:obstacle_map}
    \end{subfigure}
    \begin{subfigure}[b]{0.37\columnwidth}
        \centering
        \includegraphics[width=0.85\linewidth]{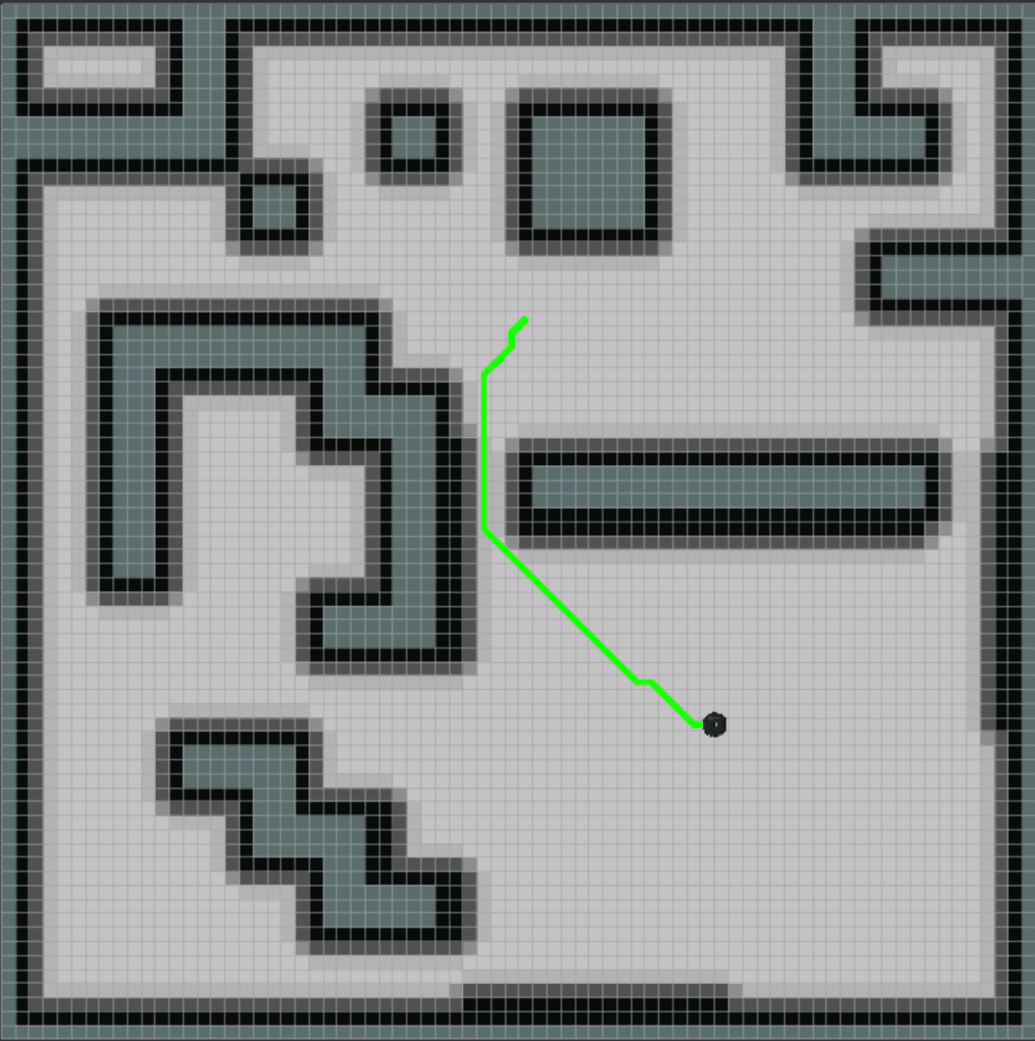}
        \caption{Initial plan}
        \label{fig:initial_plan}
    \end{subfigure}
    \vspace{0.2cm}
    
    \begin{subfigure}[b]{0.37\columnwidth}
        \centering
        \includegraphics[width=0.85\linewidth]{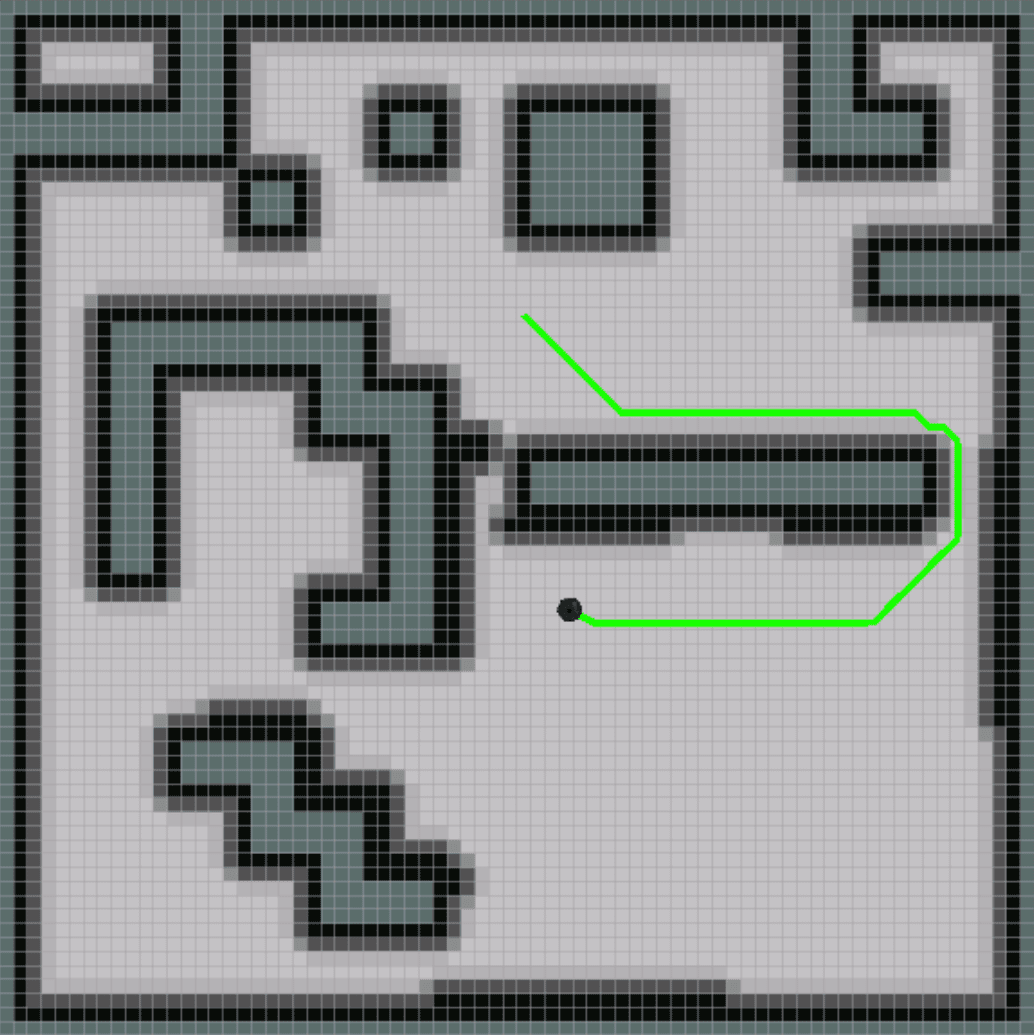}
        \caption{Re-planning - I }
        \label{fig:obstacle_1}
    \end{subfigure}
    \begin{subfigure}[b]{0.37\columnwidth}
        \centering
        \includegraphics[width=0.85\linewidth]{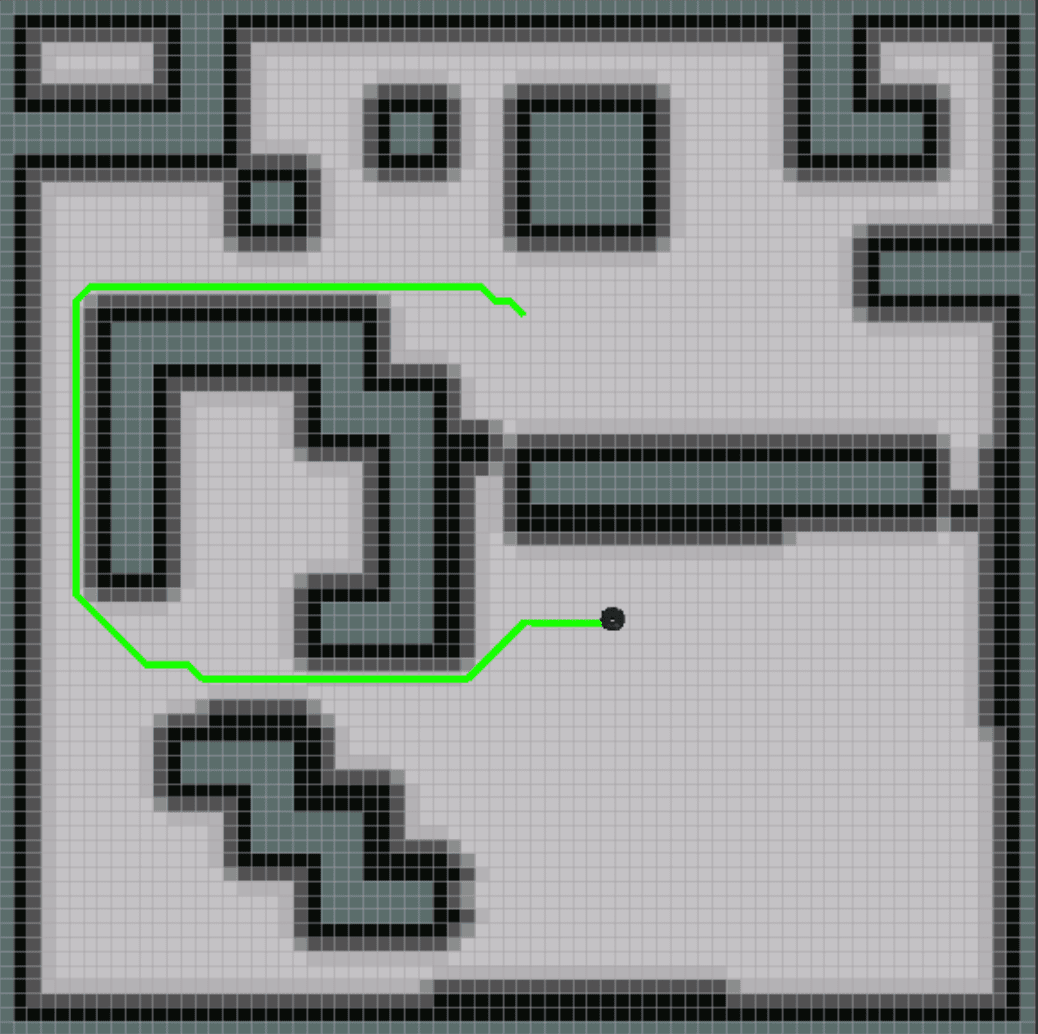}
        \caption{Re-planning - II}
        \label{fig:obstacle_2}
    \end{subfigure}
    \caption{Calculated paths for D* lite algorithm when placing dynamic obstacles}
    \label{fig:maps}
\end{figure}

To evaluate the proposed method, we have simulated using a Turtlebot3-Burger robot in a Gazebo environment consisting of square-shaped obstacles as shown in Figure \ref{fig:obstacle_map}. Dijskra and A* algorithms are used to compare the performance along with the D* lite algorithm. Initially, all algorithms were provided with the same environment with the same starting location. Two square-shaped obstacles were added at two predetermined locations while the robot is following the initially calculated global path. When adding dynamic obstacles, the same conditions were maintained for all three algorithms to obtain fair results. All results presented in these experiments are averaged over 5 runs. Figures \ref{fig:initial_plan}, \ref{fig:obstacle_1}, and \ref{fig:obstacle_2} shows the calculated paths generated by D* lite algorithm through the proposed method when adding the two obstacles.

Obtained results of this experiment are shown in Table \ref{tab:performance}, where the initial planning time column indicates the initial time taken by the algorithm to find the shortest path and the replanning time column indicates the time taken to recalculate the shortest path after adding the first and second dynamic obstacles respectively. The total time column indicates the total planning time taken by the path planner to find the shortest path after considering all dynamic changes in the environment. 

\vspace{-0.15cm}

\vspace{-0.4cm}

\begin{table}[H]
\centering
\begin{tabular}{|l|l|l|l|l|}
\hline
\multicolumn{1}{|c|}{\multirow{2}{*}{\textbf{Algorithm}}} & \multicolumn{1}{c|}{\multirow{2}{*}{\begin{tabular}[c]{@{}c@{}}Initial \\ planning \\ time (s)\end{tabular}}} & \multicolumn{2}{c|}{\begin{tabular}[c]{@{}c@{}}Replanning time \\ after \\ placing obstacle (s)\end{tabular}}                                               & \multicolumn{1}{c|}{\multirow{2}{*}{\begin{tabular}[c]{@{}c@{}}Total \\ time \\(s)\end{tabular}}} \\ \cline{3-4}
\multicolumn{1}{|c|}{}                                    & \multicolumn{1}{c|}{}                                                                                     & \multicolumn{1}{c|}{\begin{tabular}[c]{@{}c@{}}Obstacle \\ 1\end{tabular}} & \multicolumn{1}{c|}{\begin{tabular}[c]{@{}c@{}}Obstacle \\ 2\end{tabular}} & \multicolumn{1}{c|}{}                                                                       \\ \hline
Dijkstra                                                  &3.66 $\pm${0.34} & 9.09$\pm${0.44} &8.64$\pm${0.61} & 21.39$\pm${1.11}                                                                                          \\ \hline
A*                                                        &0.17$\pm${0.04}  & 2.87$\pm${0.14} & 5.28$\pm${0.13} & 8.32$\pm${0.21}                                                                                 \\ \hline
D* lite                                                   &\textbf{0.12}$\pm${0.02} &\textbf{0.39}$\pm${0.04} &\textbf{0.42}$\pm${0.08} &\textbf{0.93}$\pm${0.08} \\ \hline
\end{tabular}
\caption{Performance comparison of utilizing static and dynamic planners as global planners using the proposed approach}
\label{tab:performance}
\end{table}

D* lite algorithm which was implemented using the proposed method has taken the least planning time. Therefore, it is evident that adding a dynamic path planner as the global path planning algorithm using the proposed method can outperform static path planners used as the global path planner. It was observed that dynamic global path planning saves more than 9 times of the planning time compared to static planners used as the global planner which re-plan from scratch when the environment changes. This method of re-planning could be very efficient for applications where the surrounding environment changes very rapidly since our proposed method enables the dynamic global path planner to reuse previous search information. In the real world, even if we are aware that the environment is mostly static, we still need to call the planners at a planner frequency rate rather than planning only one time at the beginning in order to tackle sudden or minor changes in the environment. Therefore, we evaluate our algorithm by moving towards a static environment where the planners are called at the planner frequency rate. We find that our approach is still beneficial for static planners since we reduce computations as discussed in section \ref{sec:dynamic_path_planning} when the plans are requested repeatedly until the robot reaches the goal. This was evident by the analysis results summarized in Table \ref{tab:staticplanners}, where we compare the improvement introduced by our proposed Algorithm \ref{alg1} on static planners. In this setting, as previously, we keep the same start and goal positions, however with the exception of dynamic obstacles.

\begin{table}[H]
\centering
\begin{tabular}{|l|c|c|}
\hline
\multirow{2}{*}{Algorithm} & \multicolumn{2}{c|}{\begin{tabular}[c]{@{}c@{}}Total planning time (s) till the goal \\ position was reached\end{tabular}} \\ \cline{2-3} 
                           & \multicolumn{1}{l|}{Proposed method}        & \begin{tabular}[c]{@{}c@{}}Without the \\ Proposed method\end{tabular}       \\ \hline
Dijkstra                   & \textbf{14.56} $\pm \ 2.41$                                   & 37.43 $\pm \ 3.47$                                                                       \\ \hline
A*                         & \textbf{1.70} $\pm \ 0.36$                                     & 3.96 $\pm \ 0.25$                                                                       \\ \hline
\end{tabular}
\caption{Performance comparison of static planners in static environments }
\label{tab:staticplanners}
\end{table}

\vspace{-0.50cm}

\section{Limitations and Future Work}

This section explores the current limitations that exist in AlphaBot2 robotics platform and the current software system setup. Since AlphaBot2 consists of only N20 micro gear motors, the weight factor of the additional hardware must be considered while performing the modifications as we did when installing the camera adapter. Since the AlphaBot2 does not have feedback sensors such as motor encoders and IMUs, error correction control systems cannot be implemented, posing a limitation on the level of movement precision we can achieve. Furthermore, when the robot operates in areas where less features are present (Eg: plain coloured areas), there were issues in localizing the robot due to loss of visual odometry. This issue persisted when the robot performs fast movements as well. Therefore, we implemented the control system in such a way that fast movements are suppressed, especially in pure angular velocity commands.

Considering these limitations, a feedback sensor could be installed to minimize errors in movements and perform sensor fusion which enables better odometry. Due to the compactness of the base PCB, installing a motor encoder will require major changes in the PCB itself, therefore, future work could use an Intel RealSense D435i which has an inbuilt IMU or a separate IMU sensor.
Furthermore, SLAM could be explored using a LIDAR sensor on AlphaBot2 in future. For researchers to make use of our implementation, we intend to release a ROS package added with a Gazebo model of AlphaBot2 which will enable off-the-shelf simulations.

\vspace{-0.10cm}

\section{Conclusion}

In this work, ROS supported mobile robots and their importance in education and research fields were identified. 
Most of the ROS supported mobile robots which are capable of performing complex vision algorithms are very costly, therefore, it poses a barrier for a beginner to get hands-on experience with computer vision-based navigation algorithms. Considering a low-cost robotics platform - AlphaBot2, a workflow starting from modifying its hardware to configuring an end-to-end autonomous navigation system based on ROS was presented. By default, ROS does not facilitate using dynamic path planners as global planners to make use of their full potential.
Therefore, an algorithm to make use of dynamic path planners as global path planners was presented and  the performance was benchmarked using the D* lite algorithm as the dynamic global path planner, and compared it against A* and Dijkstra's algorithms.  The modified robotics platform and the proposed method were tested in both the simulation and in real-world environments to validate the functionality. The presented workflow will be beneficial for robotics researchers, students, and hobbyists to set up a low-cost robotics platform and implement and test their algorithms for autonomous navigation. 

\vspace{-0.10cm}
\section{Acknowledgements}

The authors would like to thank Mr. Paul Flick and Mr. Nicholas Panitz for their support throughout the project and Dr. Navinda Kottege, Dr. Tirthankar Bandyopadhyay, Mr. Thomas Hines, and Mr. Troy Cordie for their helpful suggestions and feedback. This work was fully funded by the Commonwealth Scientific and Industrial Research Organisation (CSIRO), Australia.

\vspace{-0.15cm}

\bibliographystyle{IEEEbib}
{\footnotesize \bibliography{references_fixed}}

\end{document}